# MIDV-2020: A Comprehensive Benchmark Dataset for Identity Document Analysis

*K.B. Bulatov [1,2], E.V. Emelianova[2,3], D.V. Tropin [2,4], N.S. Skoryukina [1,2], Y.S. Chernyshova [1,2],*
*A.V. Sheshkus [1,2], S.A. Usilin [1,2], Z. Ming [5], J.-C. Burie [5], M. M. Luqman [5], V.V. Arlazarov [1,2]*

[1] *Federal Research Center «Computer Science and Control» or Russian Academy of Sciences, Moscow, Russia*
[2] *Smart Engines Service LLC, Moscow, Russia*
[3] *National University of Science and Technology «MISiS», Moscow, Russia*
[4] *Moscow Institute of Physics and Technology (State University), Moscow, Russia*
[5] *L3i Laboratory, La Rochelle University, La Rochelle, France*

*Abstract*

Identity documents recognition is an important sub-field of document analysis, which deals with tasks of robust document detection, type identification, text fields recognition, as well as identity fraud prevention and document authenticity validation given photos, scans, or video frames of an identity document capture. Significant amount of research has been published on this topic in recent years, however a chief difficulty for such research is scarcity of datasets, due to the subject matter being protected by security requirements. A few datasets of identity documents which are available lack diversity of document types, capturing conditions, or variability of document field values. In addition, the published datasets were typically designed only for a subset of document recognition problems, not for a complex identity document analysis. In this paper, we present a dataset MIDV-2020 which consists of 1000 video clips, 2000 scanned images, and 1000 photos of 1000 unique mock identity documents, each with unique text field values and unique artificially generated faces, with rich annotation. For the presented benchmark dataset baselines are provided for such tasks as document location and identification, text fields recognition, and face detection. With 72409 annotated images in total, to the date of publication the proposed dataset is the largest publicly available identity documents dataset with variable artificially generated data, and we believe that it will prove invaluable for advancement of the field of document analysis and recognition. The dataset is available for download at ftp://smartengines.com/midv-2020 and http://l3i-share.univ-lr.fr.

<u>Keywords</u>: Document analysis, Document recognition, Identity documents, Open data, Video recognition, Document location, Text recognition, Face detection.

<u>Acknowledgements</u>: This work is partially supported by Russian Foundation for Basic Research (projects 19-29-09066 and 19-29-09092). All source images for MIDV-2020 dataset were obtained from Wikimedia Commons. Author attributions for each source images are listed in the original MIDV-500 description table (ftp://smartengines.com/midv-500/documents.pdf). Face images by Generated Photos (https://generated.photos).

## Introduction

Document analysis and recognition is a vast and growing research field, with studies covering topics of image processing, general computer vision, machine learning, intelligent systems, computational optics, and much more. Application of automated document analysis systems to a wide range of industries and processes created a need for more robust and precise methods of document identification and recognition, and thus an active demand for research in these fields.

A particular variety of document types which play a crucial role are identity documents, such as ID cards, passports, driving licences, and others. These documents are usually issued by governments, have strict design and security features, and their main goal is to define, verify and prove the holder's identity. The scope of usage of automatic system for identity document analysis include simplification and automatization of data entry when filling official forms [1], remote person identification [2], remote age checking [3], Know Your Customer/Anti Money Laundering (KYC/AML) procedures [4], and provision of governmental, financial, and other services.

The global demand for reliable automated identity document processing systems lead to the need of developing new methods for document detection, identification, and location; segmenting the document into relevant fields; extraction of text fields and graphical elements, such as signatures and faces; as well as anti-spoofing, validating the document authenticity, and detection of markers of potential fraud. The variation of input sources has also been growing: while traditional methods of scanning identity documents using flatbed or specialized scanners are still relevant, remote identification procedures now require the document analysis systems to process camera-captured photos, as well as perform ID recognition in a video stream.

One of the most important aspects of effective research of new document analysis methods is the availability of openly accessible datasets for training and benchmarking. However, when it comes to identity documents, this issue becomes very hard to resolve, since identity documents by their nature contain sensitive personal data, which is protected by worldwide regulations.

As was mentioned in the previous sections, since identity documents by their nature contain sensitive information, there are very few publicly available datasets of identity document images, and those which exist contain either partial information, or contain synthetic examples of ungenuine documents. Existing datasets dedicated specifically to identity document images include LRDE Identity Document Image Database (LRDE IDID) [5], the recently published Brazilian Identity Document Dataset (BID Dataset) [4], and the Mobile Identity Document Video dataset family (MIDV) [6, 7], to which the dataset presented in this paper also belongs. Some larger datasets, dedicated to address the issues of a broader document analysis problem, such as the ones from SmartDoc family [8], also contain identity document images.

Existing datasets of identity document images have disadvantages which present themselves when the datasets are used by researchers as benchmarks for specific identity document analysis tasks. The LRDE IDID [5] and the identity documents subset of SmartDoc [8] comprise a small amount of document samples, which allows them only to be used as reference benchmarks, without deeper analysis of identity processing methods. BID Dataset [4] addresses that issue, featuring 28800 synthetically generated document images with 8 different document types. At the same time, the images of BID Dataset were generated with artificial inscription of text field values over the automatically masked document regions, which might lead to the field presentation widely different from the one on actual documents, and the document owner faces were blurred in each document image, which make the dataset impossible to use for evaluation of face detection and location methods. In addition, BID Dataset only features ideally cropped documents, however some further datasets from the same team include synthetic scans and photographs as well.

The first dataset of the MIDV family was MIDV-500 [6], which contained 500 video clips of 50 identity documents, 10 clips per document type. The identity documents had different types, and mostly were «sample» or «specimen» documents which could be found in WikiMedia and which were distributed under public copyright licenses. The dataset focused on mobile video capture, featured clips shot using two smartphones (Apple iPhone 5 and Samsung Galaxy S3) under five distinct conditions: «Table», «Keyboard», «Hand», «Partial», and «Clutter».

The conditions represented in MIDV-500 thus had some diversity regarding the background and the positioning of the document in relation to the capturing process, however they did not include variation in lighting conditions, or significant projective distortions. To address the latter issues a dataset MIDV-2019 [7] was later published as an extension of MIDV-500, which featured video clips captured with very low lighting conditions and with higher projective distortions. Video clips of MIDV-2019 were captured using Apple iPhone XS Max and Samsung Galaxy S10. The dataset was also supplemented with photos and scanned images of the same document types [9] to represent typical input for server-side identity document analysis systems.

Since its publication, MIDV-500 dataset and its extension MIDV-2019 were used to evaluate the methods of identity document images classification [10-12]; identity document location [9, 13], including the methods based on semantic segmentation [14]; detecting of faces on images of identity documents [15]; and methods related to text fields recognition, including single text line recognition [16], per-frame recognition results combination [17, 18] and making a stopping decision in a video stream [19, 20]. The dataset was also used to evaluate the methods of choosing a single best frame in the identity document video capture [21] and assessing the quality of the frame for its processing by an identity analysis system [22], detection and masking of sensitive and private information [23] and general ID verification [24].

The main disadvantage of the datasets of the MIDV family is the scarcity of different document samples — all images were made using the same 50 physical document samples. The fact that the text field data and the graphical data (signatures, faces) are not variable, makes it harder to produce reliable evaluations of identity document analysis algorithms, or even to calculate meaningful benchmarks.

In this paper, we present a new dataset MIDV-2020, based on the 10 document types featured previously in MIDV-500 and MIDV-2019. One of the main goals of the MIDV-2020 dataset, presented in this paper, is to provide variability of the text fields, faces, and signatures, while retaining the realism of the dataset. The dataset consists of 1000 different physical documents (100 documents per type), all with unique artificially generated faces, signatures, and text fields data. Each physical document was photographed, scanned, and for each a video clip was captured using a smartphone. The ground truth includes ideal text field values, geometrical position of documents and faces in each photo, scan, and video clip frame (with 10 frames-per-second annotation). We believe that MIDV-2020 will prove invaluable for the advancement of the field of document analysis and recognition, as well as serve as a benchmark for modern identity document processing systems.

The dataset is publicly available for download at ftp://smartengines.com/midv-2020 and http://l3i-share.univ-lr.fr, the alternative mirrors have identical content.

## *1. Dataset*

In this section, we will present the basic composition of the dataset, description of its contents and annotations.

## 1.1. Composition of the dataset

The set of base document types for MIDV-2020 comprises 10 document types, each present in previously published MIDV-500 and MIDV-2019 datasets. The identity document types of MIDV-2020 are listed in Table 1, with the codes of the PRADO database [25] for each document type except the Internal passport of Russia. 100 sample documents were created for each of the 10 document types present in the dataset.

*Table 1. Document types featured in MIDV-2020*

| # | Document type code | Description | PRADO code | MIDV-500 code |
|---|---|---|---|---|
| 1 | alb_id | ID Card of Albania | ALB-BO-01001 | 01 |
| 2 | aze_passport | Passport of Azerbaijan | AZE-AO-02002 | 05 |
| 3 | esp_id | ID Card of Spain | ESP-BO-03001 | 21 |
| 4 | est_id | ID Card of Estonia | EST-BO-03001 | 22 |
| 5 | fin_id | ID Card of Finland | FIN-BO-06001 | 24 |
| 6 | grc_passport | Passport of Greece | GRC-AO-03003 | 25 |
| 7 | lva_passport | Passport of Latvia | LVA-AO-01004 | 32 |
| 8 | rus_internalpassport | Internal passport of Russia | n/a | 39 |
| 9 | srb_passport | Passport of Serbia | SRB-AO-01001 | 41 |
| 10 | svk_id | ID Card of Slovakia | SVK-BO-05001 | 42 |

To create unique sample documents the original sample images obtained from Wikimedia Commons (we used the same source images as in MIDV-500 [6]) were edited to remove non-persistent data (such as signature, photo, and text field values). Then, the generated text field values for each field were added to the images using a font which closely resembles the original font of the sample. In addition, for each document a mock signature was drawn, vaguely resembling the spelling of the last name, and a unique face picture was added.

The values of gender, birth date, issue date, and expiry date were generated in accordance with the specifics of issuing countries and a pre-set distribution of age and gender parameters:

1. 80% of generated documents correspond to adult holders (ages 18 through 60), 10% of generated documents correspond to seniors (ages 60 through 80) and 10% to children and adolescents (17 or less) depending on the minimum age for the document issue.
2. 50% of the generated documents correspond to female holders, and 50% to male holders.

Names and addresses were generated using the databases of existing names and examples of addresses available online, using Wikipedia listings of names [26], and online name generators [27].

Artificially generated face pictures by the Generated Photos service [28] were used to provide a unique face for each document. The service lists StyleGAN [29] as the approach used to create artificial face images. The images were taken either in color or in grayscale, depending on the original document sample, and were repeated if the document contained multiple copies of the face picture with opacity corresponding to the original sample. The pictures were selected to approximately match the holder age.

The template images prepared in such a way were printed to scale on a glossy photo paper, laminated, and cropped. All four corners of the card-sized documents, and the bottom two corners of the passport-sized documents were rounded using a 4mm corner rounder.

The original template images, which were used for printing, along with their annotation files are placed in the dataset within the «templates.tar» archive. There are 1000 template images in total. The images are divided according to the document type and for each document type they are numbered from «00.jpg» to «99.jpg». An example of a template image is presented in Fig. 1.

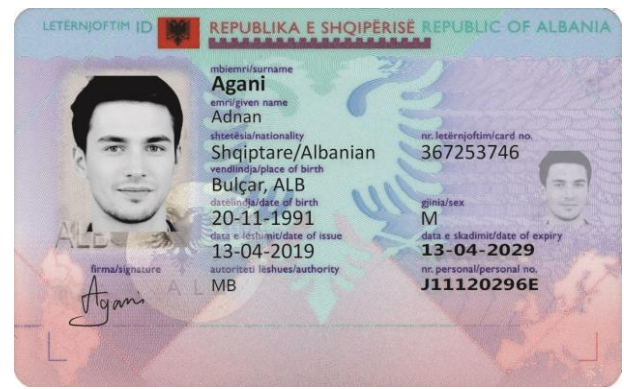

*Fig. 1. An example of a template image (image alb_id/00)*

For each image a corresponding annotation file in JSON format is provided, readable with VGG Image Annotator v2.0.11 [30], available on its website [31] (see the fragment of the annotation tool screenshot in Fig. 2). The annotation for each image includes rectangular bounding boxes of the main document holder photo (named «photo»), the bounding box of the face oval (named «face»), the bounding box of a signature field (named «signature») and the rectangles corresponding to the positions of text fields. For each text field its exact value is provided. The upper and lower boundaries of the text field rectangle correspond to the cap line and baseline respectively, and for each text field additional information is provided to indicate whether the field has lowercase letters, descenders, or ascenders. Finally, for document types where some text fields are oriented vertically, an additional orientation attribute is provided for such text fields which indicates the angle of the field's counterclockwise rotation expressed in degrees.

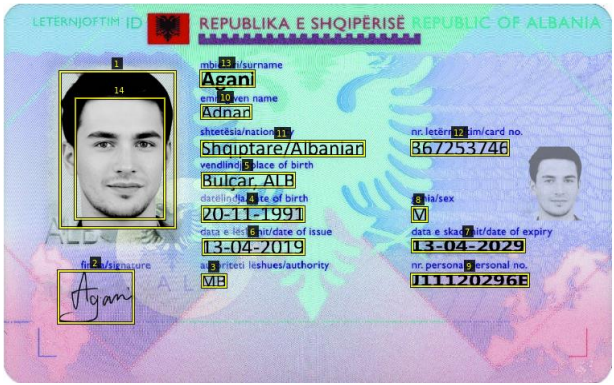

*Fig. 2. An example of a template image with fields annotation (image alb_id/00)*

### 1.2. Scans

Each physical document sample was scanned using Canon LiDE 220 and Canon LiDE 300 scanners, in two different modes. The first mode shows the document in an upright position, with a slight margin, near the top-right corner of the scanned image. The margin is enforced with a pink piece of paper. The second mode shows the document in an arbitrary place within the scanned image frame, and rotated to an arbitrary angle. All scanned images have the resolution of 2480x3507 pixels, they have not been automatically cropped. Examples of an «upright» and «rotated» scans are presented in Fig. 3.

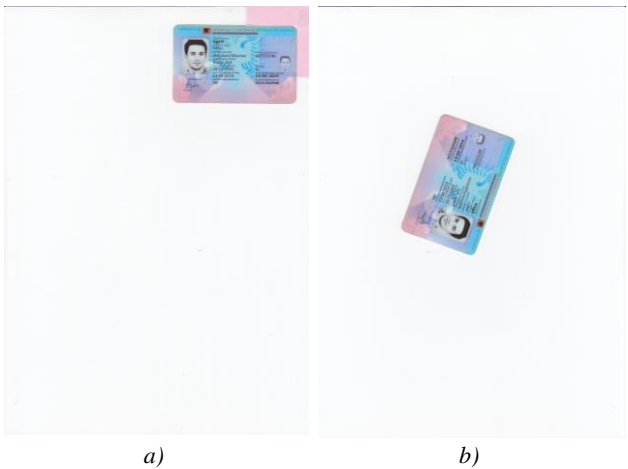

*a)                             b)*

*Fig. 3. Examples of MIDV-2020 scans: a) upright scan, b) rotated scan (image alb_id/00)*

The scanned images were captured in TIFF format, then converted to JPEG using ImageMagick 7.0.11 with default parameters. The JPEG images with their annotations are placed in the dataset within the archives «scan_upright.tar» and «scan_rotated.tar». The original TIFF images are placed within the archives «scan_upright_tif.tar» and «scan_rotated_tif.tar». There are 1000 upright scans and 1000 rotated scans in total. The names of the scanned images correspond to the names of the template image from which the physical document was created.

The annotations for scanned images are provided in JSON format readable with VGG Image Annotator v2.0.11 and feature the bounding boxes of the face oval (marked with a field name «face») and the document quadrangle (marked with a field name «doc_quad»). The first vertex of the quadrangle always corresponds to the top-left corner of the physical document, and the rest of the vertices are provided in a clockwise order.

### 1.3. Photos

For each physical document sample a photo was taken, given various conditions and two smartphones. Half of the photos were captured using Apple iPhone XR, and the other half using Samsung S10. The following capturing conditions were used:

1. Low lighting conditions (20 documents of each type);
2. Keyboard as a background (10 documents of each type);
3. Natural lighting, captured outdoors (10 documents of each type);
4. Table as a background (10 documents of each type);
5. Cloth with various textures as a background (10 documents of each type);
6. Text document as a background (10 documents of each type);
7. High projective distortions of the document (20 documents of each type);
8. Highlight from the sun or lamp hides a portion of the document (10 documents of each type).

Examples of photos captured with different conditions are presented in Fig. 4.

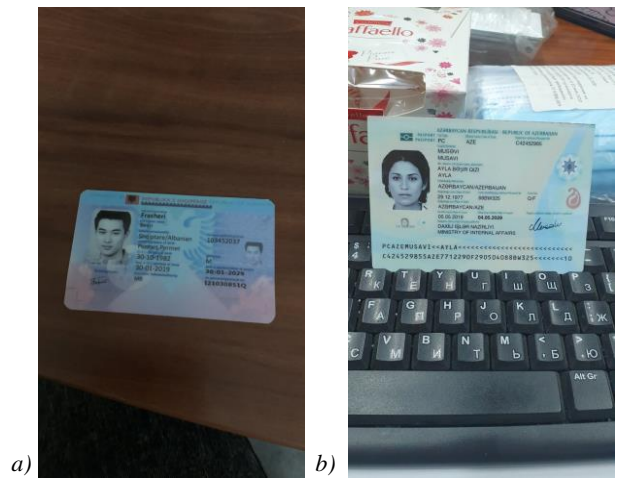

*a)                             b)*

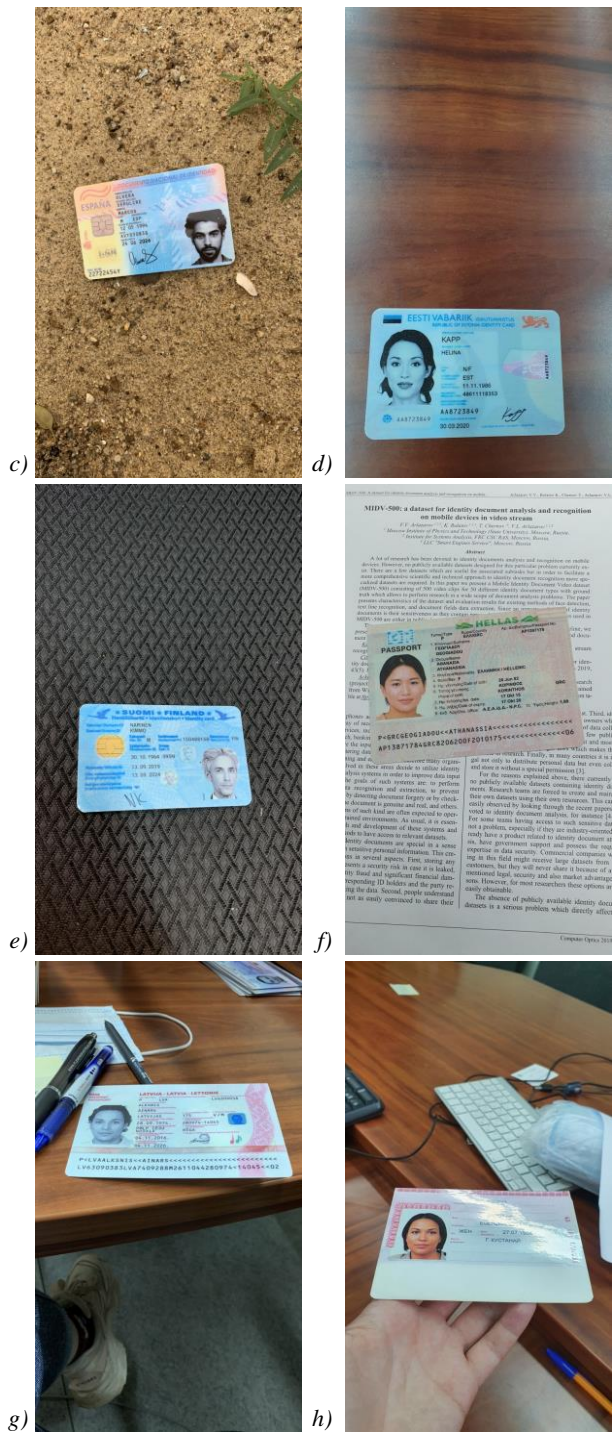

*Fig. 4. Examples of MIDV-2020 photos: a) low lighting (image alb_id/70); b) Keyboard (image aze_passport/35); c) Outdoors (image esp_id/40); d) Table (image est_id/55); e) Cloth (image fin_id/90); f) Text background (image grc_passport/25); g) Projective (image lva_passport/00); h) Highlights (image rus_internalpassport/65)*

The ranges of codes of the document samples which correspond to each of the listed conditions are specified in Table 2.

*Table 2. Ranges of image codes corresponding to photo capturing conditions*

| Capturing conditions and smartphone models | Samsung S10 | Apple iPhone XR |
|---|---|---|
| Low lighting | 80-89 | 70-79 |
| Keyboard in the background | 35-39 | 30-34 |
| Natural lighting, outdoors | 45-49 | 40-44 |
| Table in the background | 55-59 | 50-54 |
| Cloth in the background | 95-99 | 90-94 |
| Text documents in the background | 25-29 | 20-24 |
| Projective distortions | 10-19 | 00-09 |
| Highlight present | 65-69 | 60-64 |

All photos are stored in JPEG format and have the resolution of 2268x4032 pixels. The images with annotations are placed in the dataset within the «photo.tar» archive. The names of the images correspond to the names of the template from which the physical document was created. There are 1000 photos in total. The annotations format is the same as for the scans.

### 1.4. Video clips

For each document sample a video clip was captured, using the similar distribution of the capturing conditions as with photos. Each clip was captured vertically, in a resolution of 2160x3840 pixels, with 60 frames per second. There are 1000 clips in total, each with different lengths. The numbers of the document samples which correspond to each of the capturing conditions are specified in Table 3.

*Table 3. Ranges of clip codes corresponding to video capturing conditions*

| Capturing conditions and smartphone models | Samsung S10 | Apple iPhone XR |
|---|---|---|
| Low lighting | 00-09 | 10-19 |
| Keyboard in the background | 20-24 | 25-29 |
| Natural lighting, outdoors | 60-64 | 65-69 |
| Table in the background | 30-34 | 35-39 |
| Cloth in the background | 40-44 | 45-49 |
| Text documents in the background | 50-54 | 55-59 |
| Projective distortions | 70-79 | 80-89 |
| Highlight present | 90-94 | 95-99 |

The original captured clips were stripped of audio channel and separated into frames using ffmpeg version n4.4 with default parameters. To provide a rich annotation each video clip was annotated with 10 frames per second. To this end, each 6-th frame of each clip was retained (thus, the dataset only retained the separate frames «000001.jpg», «000007.jpg», «000013.jpg», etc.). The smallest clip has 38 frames, the largest has 129 frames. Overall, the dataset includes 68409 annotated video frames.

Frames with the corresponding annotations are placed in the dataset within the «clips.tar» archive. The video files are located in the «clips_video.tar» archive. The annotation of each frame has the same format as for the photos and scanned images.

## 2. Benchmarks

While the main goal of the paper is to present an identity document dataset MIDV-2020, in order to provide a baseline for future research involving the dataset, in the following sections several benchmarks using MIDV-2020 will be presented. Based on the topics for which the datasets of the MIDV family were used (see the Introduction section), the benchmark will include the task of content-independent document boundaries location, semantic segmentation of the document body from the background, feature-based document location and type identification, text fields recognition, and face detection.

### 2.1. Content-independent document boundaries location

Detection and location of a document in an image or a video frame is one of the first major steps in an identity document analysis system. When the set of possible document types is not known in advance, or if the knowledge of the general structure of the document content is not available, the systems have to resolve to content-independent methods for document location. One of the types of such methods is locating document boundaries, under the assumption that the document is a planar rectangular object with a known aspect ratio (or a known set of possible aspect ratios).

Consider an image which contains an identity document. Let us assume that the document is a planar rigid rectangle with a known aspect ratio. The corners of the rectangle could be either sharp or rounded. The content of the document is not known in advance, however we can assume that the content color is distinguishable from the background. The image is captured using a camera or a scanner. The capturing conditions are uncontrolled: the background, lighting, and document orientation can be arbitrary, the document may partially be outside the frame or occluded by another object. We assume that in the image there is only one document with a given aspect ratio. The task is to estimate the coordinates of the vertices of the quadrangle corresponding to the document boundary, down to their convex re-enumeration. An input to the algorithm, besides the image $I$, comprises the size of the document template (henceforward, we will use $t$ as a rectangle with vertices in points *(0; 0), (w; 0), (w; h), (0; h)*, where $w$ and $h$ are linear sizes of the template). Based on this input the algorithm needs to estimate the quadrangle $q$ which would satisfy a binary quality criterion $L(q, t, m)$, where $m$ represents a ground truth quadrangle.

Binary quality criterion $L$ may be based on (i) a Jaccard score:

$$\text{IoU}(q,m) = \frac{\text{area}(q \cap m)}{\text{area}(q \cup m)}; \quad (1)$$

(ii) Jaccard score calculated in the coordinate system of a ground truth quadrangle $m$:

$$\text{IoU}^{gt}(q,m,t) = \frac{\text{area}(Mq \cap t)}{\text{area}(Mq \cup t)}, \quad (2)$$

where $M$ designates a homography such that $Mm = t$;
(iii) maximum distance between the corresponding vertices of $q$ and $m$ computed in the coordinate system of $q$:

$$D^0(q,m,t) = \max_i \frac{\|t_i - Hm_i\|_2}{P(t)}, \quad (3)$$

where $H$ represent a homography such that $Hq = t$ and $P(t)$ is a perimeter of the template. Since the quadrangle of the document should be detected down to their convex re-enumeration of its vertices the final binary quality criterion based on (3) is a follows:

$$D^4(q,m,t) = \min_{q^{(i)} \in Q} D^0(q^{(i)},m,t), \quad (4)$$

where $Q$ is the set of convex vertex re-enumerations $\{[a,b,c,d];[b,c,d,a];[c,d,a,b];[d,a,b,c]\}$.

There exist three base approaches for detecting a quadrangle of the document boundaries: corners detection [32, 33], straight lines detection and analysis [34-37], and the analysis of salient regions [2, 38]. As shown in [35], the representatives of all three classes successfully solve the problem given a subset of an open dataset SmartDoc [39], which complies to the problem statement.

There is an openly available model from [32], which uses a convolutional neural network for document vertex detection. For the purposes of a baseline evaluation, we used a model trained using a dataset of A4-size documents (SmartDoc subset and own-collected). We also asked the authors of [35-37] to evaluate their algorithms using MIDV-2020. The method described in [35] uses straight lines (for generating quadrangle hypotheses) and salient regions (for comparing and ranging the hypotheses). This algorithm shows state-of-the-art results on the open dataset MIDV-500 [6] and can be used on low-end computational devices. The algorithms [35] and [37] both require the knowledge of the document's aspect ratio and the knowledge of the camera internal parameters (focal length and position of a principal point). Although we do not know the latter from our problem statement, in experiments with [35] and [37] we used the focal length equal to 0.705 from the diagonal of the image and the position of principal point in the center of the image. The algorithm [36] as well as [35] uses straight lines and salient regions however does not require knowledge of the aspect ratio and internal parameters of the camera.

In order to estimate the accuracy given a dataset of images we will use the following statistics: the rate of correct answers for the two binary quality criteria: IoU > 0.9 (see Table 4) and $D^4 < 0.02$ (see Table 5), as well as a mean Jaccard score $\text{IoU}^{gt}$ (see Table 6). The source code of the statistics implementation published at

https://github.com/SmartEngines/hough_document_localization, as referenced in [35].

*Table 4. Document location performance (percentage of images with IoU > 0.9) of content-independent methods on MIDV-2020*

| Method | MIDV-2020 subset | | | |
|---|---|---|---|---|
| | Upright scans | Rotated scans | Photos | Video frames |
| [32] | 0.00 | 2.50 | 34.00 | 23.34 |
| [35] | **100.00** | **91.30** | **87.00** | **85.55** |
| [36] | 72.80 | 50.80 | 82.00 | 78.01 |
| [37] | 98.20 | 91.20 | 83.30 | 81.74 |

*Table 5. Document location performance (percentage of images with $D^4 < 0.02$) of content-independent methods on MIDV-2020*

| Method | MIDV-2020 subset | | | |
|---|---|---|---|---|
| | Upright scans | Rotated scans | Photos | Video frames |
| [32] | 0.00 | 1.70 | 33.50 | 22.99 |
| [35] | **99.70** | **90.80** | **85.40** | **83.08** |
| [36] | 72.60 | 49.70 | 79.20 | 74.74 |
| [37] | 97.00 | 90.70 | 81.80 | 79.65 |

*Table 6. Document location performance (mean value of $IoU^{gt}$) of content-independent methods on MIDV-2020*

| Method | MIDV-2020 subset | | | |
|---|---|---|---|---|
| | Upright scans | Rotated scans | Photos | Video frames |
| [32] | 0.2142 | 0.5462 | 0.6569 | 0.6144 |
| [35] | **0.9874** | **0.9320** | **0.8977** | **0.8997** |
| [36] | 0.7353 | 0.5381 | 0.8682 | 0.8644 |
| [37] | 0.9790 | 0.9192 | 0.8797 | 0.8791 |

Example outputs of the evaluated systems are presented in Fig. 5. Note that on the Fig. 5*a* the output of the [35] and [37] totally overlap, the same on the Fig. 5*c*, but for systems [35-37].

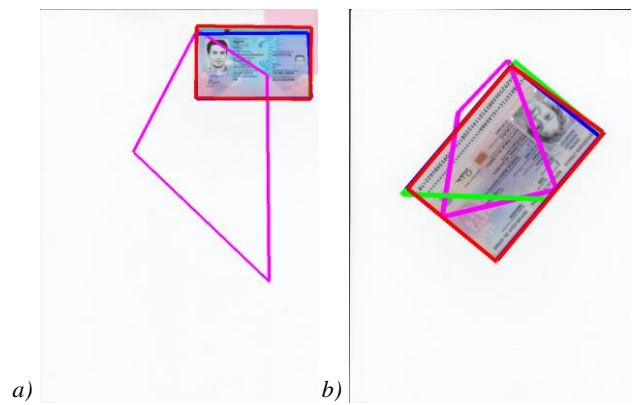

*a)                                b)*

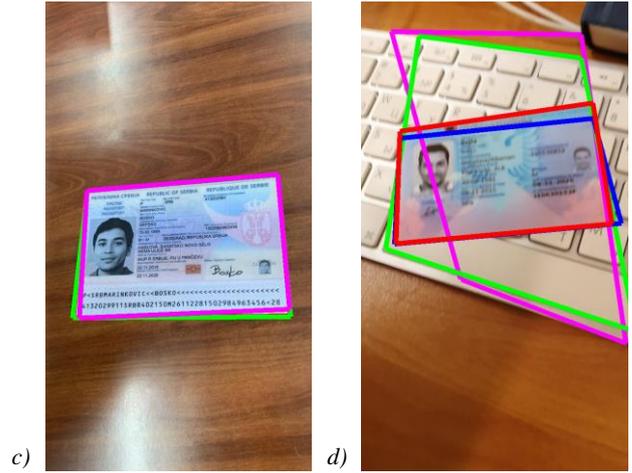

*c)                                d)*

*Fig. 5. Examples of document quadrangle detection in images from different MIDV-2020 subsets: a) Upright scans (image alb_id/00), b) Rotated scans (image srb_passport/86), c) Photo (image srb_passport/51), d) Video clip frame (frame alb_id/22/000217). Color denotes different methods: [32] — magenta, [35] — red, [36] — green, [37] — blue*

The results of the system presented in [32], while being quite low, correspond to the data presented in the original paper: the mean value of $IoU^{gt}$ on the 5-th subset of the SmartDoc dataset (excluding the training subset) was reported to be 0.67. Lower results on the 5-th subset of SmartDoc was explained by the presence of additional objects near the target document. In the case of MIDV-2020 additional complications include (i) the presence of scanned images, where the documents are shifted to the top-right corner and have small relative size; and (ii) the documents of MIDV-2020 have rounded corners, which negatively influence the second stage of the algorithm presented in [32], where the corners are recursively refined. An example of the rounded corners effect is presented in Fig. 5*c*: the system of [32] successfully detected sharp corners, but incorrectly detected the rounded ones.

As Tables 5 and 6 show, the metrics of the mean value of $IoU^{gt}$ may be insufficient for the evaluation and benchmarking of document location in context of ID recognition systems: the results of [32] on the «Rotated scans» subset is higher than that of [36], whereas the values of ($IoU > 0.9$) and ($D^4 < 0.02$) for this method amount only to a few percent, being significantly lower than [36] on the same subset.

### 2.2. Document location via semantic segmentation

An alternative approach to the search of document boundaries is performing semantic segmentation of the image to determine which pixels correspond to the document and which to the background. Here we can assume that the document is pseudo-rigid, possibly warped quadrangle and unknown internal content. The task is to classify each pixel to either document or a background. The most natural metrics to evaluate such a task is to compute a pixel-wise Jaccard score [13].

To evaluate the semantic segmentation method, we used the published methods HU-PageScan [13] and HoughEncoder [14], both based on a U-net modification. HU-PageScan has an openly available source code and a pre-trained model, which was used for this evaluation. We asked the authors of HoughEncoder to evaluate their model with training on MIDV-500 [6] and MIDV-2019 [7]. The results of the evaluation are presented in Table 7.

*Table 7. Document location performance (mean value of IoU) of semantic segmentation methods on MIDV-2020*

| Method | MIDV-2020 subset | | | |
| --- | --- | --- | --- | --- |
| | Upright scans | Rotated scans | Photos | Video frames |
| [13] | **0.0941** | 0.0564 | 0.2680 | 0.1509 |
| [14] | 0.0850 | **0.1688** | **0.6508** | **0.6792** |

The results of [13] were obtained using a published source code at https://github.com/ricardobnjunior/HU-PageScan, referenced in the original paper. Examples of the semantic segmentation results on images from various MIDV-2020 subsets are presented in Fig. 6.

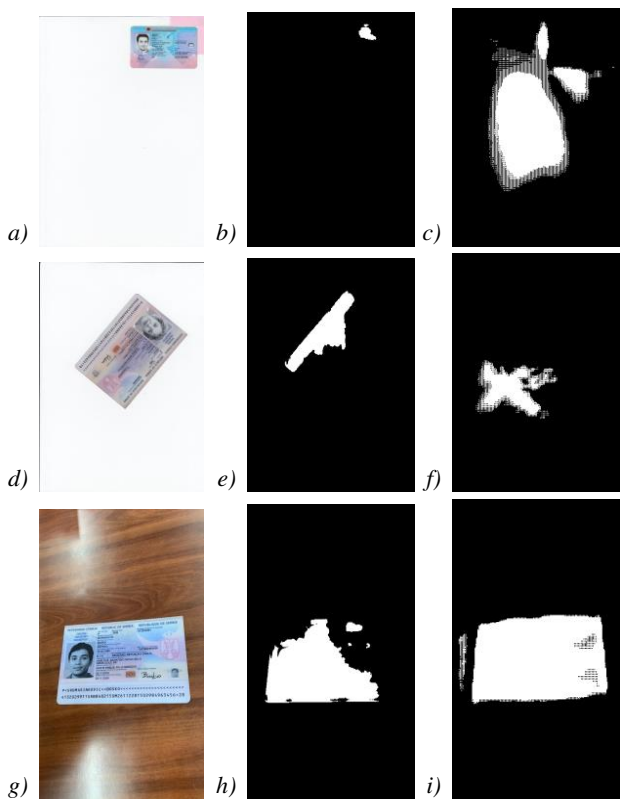

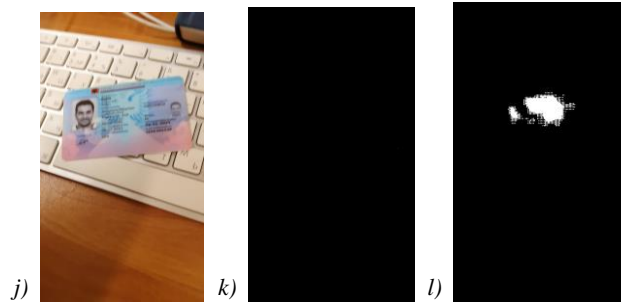

*Fig. 6. Examples of semantic segmentation results of images from MIDV-2020 dataset (the source images correspond to the ones presented in Fig. 5. a) source image; b) result of [13]; c) result of [14]; d) source image; e) result of [13]; f) result of [14]; g) source image; h) result of [13]; i) result of [14]; j) source image; k) result of [13]; l) result of [14]*

Low results of the HoughEncoder method [14] on the «Upright scans» and «Rotated scans» subsets of MIDV-2020 can be explained by its training on MIDV-500 and MIDV-2019 datasets, which do not feature scanned images.

Since both methods are trained and inferred using full images, including document contents and backgrounds, they are not, strictly speaking, content-independent. This might sometimes lead to segmenting only specific parts of the documents, for example, in Fig. 6*e* and 6*h* it can be seen that only the zones of the documents with visible text lines are classified as belonging to a document. The low baseline results for semantic segmentation-based document location methods as evaluated on MIDV-2020 (even if trained on MIDV-500 and MIDV-2019) show the need to continue the research and improve the existing methods for them to attain practical application for identity image analysis.

### 2.3. Feature-based document location and identification

The previous two sections considered document location methods which did not directly rely on the internal document content (like the boundary analysis methods), or which learned the differences between document content and background (like in the case of semantic segmentation) given a train dataset. Identity documents in general can be classified as semi-structured objects — their content consists of static elements of the document template and variable personal data. Within a single document template type the exact or approximate position of the variable personal data is known. Let us consider the task of document type identification given a scanned or photographed image. Given a predefined set of identity document templates (which, in a general case, may contain hundreds or thousands of different templates) the task is to classify the image to one of the classes representing a template, or to a null class, which correspond to the absence of a known document type. The state-of-the-art approach [40] for solving such a problem is based on local features matching. The typical base features are keypoints and their descriptors, however the feature space can be expanded with line segments, quadrangles, vanishing points, etc. [9, 41].

To provide a baseline for evaluation of such methods using MIDV-2020 dataset we used the algorithm described in [42]:

1. For each ideal document template an ideal image is used to mark up the zones with variable personal data.
2. Keypoints and their descriptors are extracted in an ideal document template image and in the analyzed image. The keypoints inside the variable data zones are discarded.
3. A set of direct matches is constructed. For each local descriptor of the analyzed image the closest descriptor from the set of all stored ideal image descriptors is found. The document types are ranged by the total number of the closest match occurrences with respect to all keypoints in the analyzed image.
4. Among the found matches the symmetric ones are selected (i.e., the ones which are the closest both for the analyzed image and for the ideal template image), and with a dominant orientation (the angle of rotation around the keypoint).
5. Using the selected set of keypoint pairs a projective transformation is estimated using RANSAC. If the projective transformation is estimated and well-defined, the number of inlier pairs is calculated for it (considered zero otherwise).
6. The result is the document template type with the highest number of inlier pairs.

Such an approach allows not only to identify the type of the document, but also to precisely locate the coordinates of the document vertices using the found projective transformation matrix.

Within the scope of the benchmark evaluation, in order to maximize reproducibility, the search for direct and inverse matching was performed without any randomized approximation, and no additional geometric restrictions were imposed, following the procedure described in [11]. For keypoints detection the SURF method was employed [43], and for matching the keypoints we evaluated the SURF descriptor [43], and the binary descriptor BEBLID [44] in 256- and 512-dimension variants. The accuracy of document type identification as measured on MIDV-2020 dataset is presented in Table 8.

*Table 8. Feature-based document identification accuracy*

| Descriptor | MIDV-2020 subset | | | |
|---|---|---|---|---|
| | Upright scans | Rotated scans | Photos | Video frames |
| SURF | **100.00** | **100.00** | 95.10 | 64.38 |
| BEBLID-256 | **100.00** | 99.90 | 98.20 | 81.75 |
| BEBLID-512 | **100.00** | **100.00** | 98.70 | **84.48** |

For an end-to-end system it is important not only to correctly identify the document type, but to correctly find the coordinates of the document's vertices as well, especially if the procedures for text fields extraction and recognition depend on a preliminary rectification of the document page. Table 9 presents the benchmark results for correct type identification and correct location, across the images presented in the MIDV-2020 dataset, using the feature-based approach. Document location quality is estimated using the $D^4 < 0.02$ criterion, introduced in section 2.1.

*Table 9. Feature-based document identification and location ($D^4 < 0.02$)*

| Descriptor | MIDV-2020 subset | | | |
|---|---|---|---|---|
| | Upright scans | Rotated scans | Photos | Video frames |
| SURF | 98.00 | 96.70 | 89.00 | 54.50 |
| BEBLID-256 | **99.00** | 97.70 | 93.10 | 72.37 |
| BEBLID-512 | 98.80 | **99.00** | **95.70** | **75.13** |

Fig. 7 shows an ideal template image with template keypoints, and in Fig. 8 the quadrangle location results of the evaluated methods are presented.

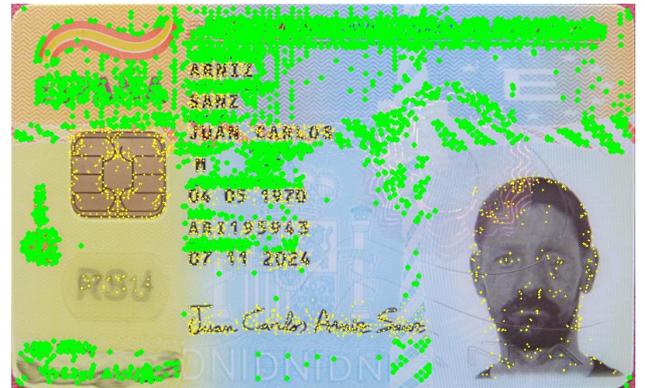

*Fig. 7. Ideal document template (from MIDV-500). Yellow points represent the keypoints in variable data zones, which are not used for document location and identification. Green points represent static element keypoints, their neighborhoods are used for descriptors computation.*

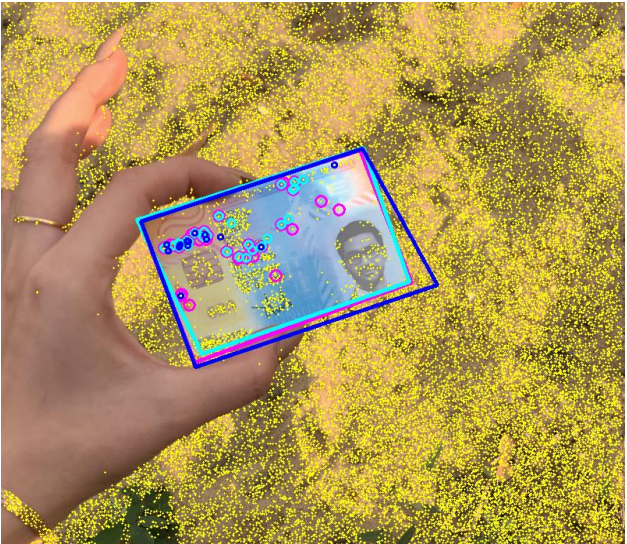

*Fig. 8. Analyzed image (photo esp_id/44). Yellow points represent all detected keypoints. Document location results and keypoint inliers are shown in color (SURF — blue, BEBLID-256 — cyan, BEBLID-512 — magenta).*

Low accuracy of the feature-based document location and identification on the video frames of MIDV-2020 in comparison with photos and scans could be due to the effects of defocus and motion blur on the ability to distinguish between local features around small static text elements. The differences between the measured accuracy for different descriptors correlate with the local patch matching accuracy measured in [44].

### 2.4. Text fields recognition

One of the most important tasks related to identity document analysis is the recognition of text fields, which are used for person authentication, identity verification, and automated form filling. In a focused text field recognition scenario, we assume that the preliminary identity document analysis steps, such as detecting and locating the document in an image, and segmenting the document into the separate fields, had already been performed. To provide a baseline for text fields recognition, we employed Tesseract OCR 4.1.1 [45, 46] that provides LSTM-based recognizer for a variety of languages. Previous studies [16] show that Tesseract OCR produces reasonable results in comparison with other popular baseline systems, e.g., Abbyy FineReader [47].

To acquire text field images without spatial distortions introduced by the previous document processing steps (document detection, per-field segmentation, etc.), we used the coordinates provided in the dataset ground truth. To obtain rectified field images we extended the annotated bounding boxes by 20% in each direction. If the field contained letters that cross the baseline or the cap line (this information is provided in the ground truth as additional attributes of each annotated field), we applied an additional 10% padding.

We divided all fields in the documents into four groups — numerical fields, fields written using numerals and letters of the English alphabet (ISO basic Latin alphabet), fields written using numerals and local alphabets, and machine-readable zones (MRZ). We excluded four fields («authority», «code», «nationality», and «birth_place_eng») of the passport of Greece from our experiment as they contain bilingual (Greek and English) text. The separation of MIDV-2020 into the field groups is specified in Table 1 of the Appendix.

For each field group, we calculated per-string recognition rate (PSR) and per-character recognition rate (PCR) [16], defined as follows:

$$\text{PSR} = \frac{L_{\text{correct}}}{L_{\text{total}}}, \qquad (5)$$

where $L_{\text{total}}$ denotes the total number of text line images, and $L_{\text{correct}}$ stands for the number of correctly recognized text lines;

$$\text{PCR} = 1 - \frac{\sum_{i=1}^{L_{\text{total}}} \min\left(\text{lev}(l_{i_{\text{ideal}}}, l_{i_{\text{recog}}}), \text{len}(l_{i_{\text{ideal}}})\right)}{\sum_{i=1}^{L_{\text{total}}} \text{len}(l_{i_{\text{ideal}}})}, \qquad (6)$$

where $\text{len}(l_{i_{\text{ideal}}})$ denotes the length of the $i$-th text line, and $\text{lev}(l_{i_{\text{ideal}}}, l_{i_{\text{recog}}})$ is the Levenshtein distance between the recognized text and the ideal text in the annotations.

For the evaluation we did not modify the Tesseract recognition results and annotations in any way, except for the conversion of multiple consecutive spaces into a single space character. The obtained evaluation results are presented in Table 10.

*Table 10. Per-field recognition accuracy (Tesseract v4.1.1)*

| Document code | Field groups | | | | | | | |
| --- | --- | --- | --- | --- | --- | --- | --- | --- |
| | Latin alphabet | | Numerical fields | | Local alphabets | | MRZ | |
| | PSR | PCR | PSR | PCR | PSR | PCR | PSR | PCR |
| alb_id | 27.91 | 62.57 | 46.73 | 70.10 | 20.80 | 56.07 | n/a | n/a |
| aze_passport | 37.57 | 64.85 | 22.98 | 46.19 | 18.78 | 54.63 | 4.51 | 51.34 |
| esp_id | 28.13 | 54.97 | 46.31 | 68.94 | 51.72 | 68.68 | n/a | n/a |
| est_id | 34.64 | 57.36 | 35.67 | 68.64 | 77.16 | 87.46 | n/a | n/a |
| fin_id | 50.83 | 54.97 | 14.73 | 55.99 | 57.36 | 73.83 | n/a | n/a |
| grc_passport | 47.52 | 70.76 | 38.71 | 57.86 | 49.91 | 68.97 | 9.64 | 64.29 |
| lva_passport | 35.89 | 60.25 | 39.26 | 67.32 | 33.84 | 62.10 | 12.61 | 62.69 |
| rus_internalpassport | n/a | n/a | 49.54 | 82.78 | 43.67 | 70.74 | n/a | n/a |
| srb_passport | 36.99 | 49.80 | 21.91 | 44.29 | 22.38 | 57.40 | 15.28 | 58.46 |
| svk_id | 68.42 | 85.89 | 8.86 | 51.18 | 50.35 | 76.79 | n/a | n/a |
| **All documents** | **39.44** | **63.72** | **29.19** | **59.31** | **37.53** | **62.49** | **10.49** | **59.22** |

The results in Table 10 correlate with previously obtained results [16] for MIDV-500. The results for MRZ are slightly better than those in [16]. In our view, it is connected with the improvement of Tesseract OCR, since we used version 4.1.1, and the authors of [16] used version 4.0.0. The other field groups cannot be so easily compared as we divide the fields in a different way, e.g., we take all the fields printed with Latin alphabet into the corresponding group, and in [16] only names and last names were included. But we still can say that MIDV-2020 presents more possibilities for text line recognition algorithms improvement, as (i) for 9 out of 10 documents the results in the Latin group are lower than results for the Latin names group in [16]; and (ii) in this paper we presented baseline results for local alphabets, which were not provided before.

### 2.5. Face detection

In the past two decades, the advancement of technology in electronics and computer science allows the widespread deployment of biometric systems in real-life applications, such as online payment, smartphone-based authentication, biometric passport etc. The face recognition-based biometric systems authorize the legitimate access of the clients by comparing an individual's live face to the face image found in his/her identity document. Consequently, counterfeiting of identity documents has also become a main form to attack face recognition-based biometric systems. Thus, either for the face recognition-based authentication with the identity document or the detection of the counterfeits, the face detection/recognition are indispensable. Moreover, face detection is the first step to start the whole processing.

Instead of detecting live faces under spontaneous scenarios as in [48, 49], detecting face on identity documents is a more challenging problem due to the conditions under which the images of identity documents have been captured being more complex and with much less constraints. The face detection on identity documents encounters more challenges such as partial face, defocus, reflection, the varying shift in space, unconstrained shooting angle and the varying illumination conditions as shown in Fig. 9.

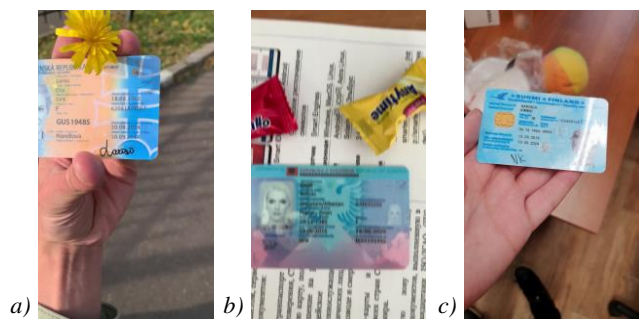

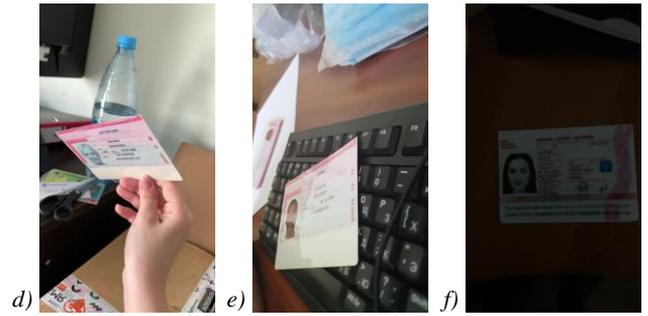

*Fig. 9. Challenging conditions for capturing the identity documents, as shown in MIDV-2020 dataset: a) partial face (frame svk_id/66/000145); b) defocus (frame alb_id/55/000019); c) reflection (frame fin_id/90/000253); d) shift in space (frame rus_internalpassport/87/000301); e) unconstrained shooting angle (frame rus_internalpassport/70/000133); and f) bad illumination (frame lva_passport/19/000103)*

Thanks to the great advances in deep neural network architectures, face detection in general face images has made great progress in the past decade [50-59]. FDDB [48] and WiderFace [49] are the most widely used datasets for training and evaluating face detection algorithms. However, FDDB and WiderFace only include the general/liveness face images in the spontaneous scenarios having no face images from the identity documents. Although FDDB and WiderFace contain the faces captured under the «unconstrained» environment, the face images on the identity documents included in the proposed MIDV-2020 have more variations as shown in Figure 9. Besides, the resolution of images in FDDB and WiderFace are relatively low (from 40x40 to 400x400 maximum) compared to the images in MIDV-2020 (2160x3840). Thus, not only for the research of face detection on identity documents but also for the research of the generalization of face detection algorithms, the MIDV-2020 is an essential supplement for the current face detection datasets.

Benefiting from the blooming of the deep learning-based generic object detection, face detection makes great progress on the extreme and real variation problem including scale, pose, occlusion, expression, makeup, illumination, blur etc. [53]. Cascade-CNN [54] firstly proposed the deep Cascade-CNNs architecture for face detection. Based on the deep cascaded multi-task framework, MTCNN (MultiTask Cascaded Convolutional Neural Networks) [55] achieved the state-of-the-art performance at that moment, which attained very high speed for face detection and alignment. PCN (Progressive Calibration Network) [56] has tackled the detection of arbitrary rotation-in-plane faces in a coarse-to-fine manner. Based on SSD (Single Shot multibox Detector) [57], $S^3FD$ (Single Shot Scale-invariant Face Detector) [58] proposed a scale compensation anchor matching strategy aiming to detect the small faces. RetinaFace [59] employed deformable context modules and additional landmark annotations to improve the performance of face detection. TinaFace [53] achieved impressive results on unconstrained face detection by revising a generic object detector. However, nowadays the face detection methods seem to make the overall algorithms and systems become more and more complex. In order to provide a strong but simple baseline method for face detection in the images of identity documents, MTCNN is selected in this work as the face detection method. MTCNN is almost the most widely used face detection method of today given its high stability, low latency and simplicity.

The evaluation of face detection on MIDV-2020 has been conducted on the five different types of images of identity documents respectively. The ground truth of bounding box of faces in MIDV-2020 has been semi-automatically annotated, i.e., the faces in the dataset have been firstly detected automatically by pretrained MTCNN and then the detected bounding boxes have been calibrated manually to serve as the ground truth of the bounding boxes of faces in MIDV-2020. In particular, there are more than one face in some identity documents: the principal face with a large scale and the other watermark-like tiny faces as shown in Figure 1. In this work, we only adopted the principal face to employ the evaluation. The MTCNN model used in this work has been pretrained on WiderFace. Thus, we can evaluate the generalization capacity of a face detection model trained on the general images performing on the heterogeneous document images. The evaluation has been conducted on one GPU (Nvidia GeForce RTX 2080).

As being an application of generic object detection, the face detection evaluation also follows the evaluation protocol of generic object detection. The widely used detection evaluation metrics Average Precision (AP) [60] is adopted to evaluate the performance of MTCNN on document images. Instead of having several different classes of objects to detect, face detection has only one class to detect. Thus, the AP here is the average precision of the detection of different faces. As same as in [60], three types of AP with different Intersection over Union (IoU) values have been used as three types of metrics for evaluating the face detection method. The metric with three different IoU thresholds indicates three different levels to locate the faces in images. The IoU threshold from 0.3 to 0.7 indicates the localization requirement from easy to hard. That means AP with IoU of 0.7 reward detectors with better localization than AP with IoU of 0.5. The source code for the evaluation scripts are available at https://github.com/hengxyz/MIDV_2020_det.

Table 11 shows the evaluation results of face detection on MIDV-2020 by MTCNN. Generally speaking, MTCNN has obtained a good detection result in every type of identity document images. Only for the scan copies of rotated identity cards, the performance of MTCNN is inferior to the others especially when IoU threshold is 0.7 since MTCNN is not designed for detecting the rotated face as PCN. Nevertheless, MTCNN still shows a comparable performance for detecting rotated faces when the localization requirement is less strict such as the AP with IoU threshold of 0.5 or 0.3. We can see that even MTCNN is pretrained on the general face images in WiderFace, it can still well generalize on the heterogeneous document images.

*Table 11. The face detection performance of MTCNN on MIDV-2020*

| MIDV-2020 subset | Average Precision (AP) | | |
|---|---|---|---|
| | Easy (IoU > 0.3) | Medium (IoU > 0.5) | Hard (IoU > 0.7) |
| Templates | 0.99899 | 0.99899 | 0.99899 |
| Upright scans | 1.00000 | 1.00000 | 1.00000 |
| Rotated scans | 0.99057 | 0.99057 | 0.90178 |
| Photos | 0.99403 | 0.99303 | 0.98865 |
| Video frames | 0.98731 | 0.98657 | 0.96160 |

Figure 10 and Figure 11 shows the ROC curves and Precision-Recall curve of face detection on different types of documents under different localization requirements, i.e., with different IoU thresholds.

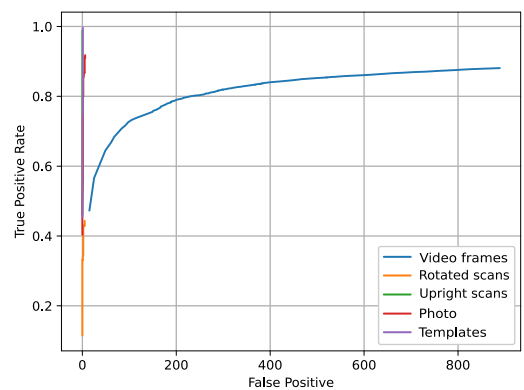
*a)*

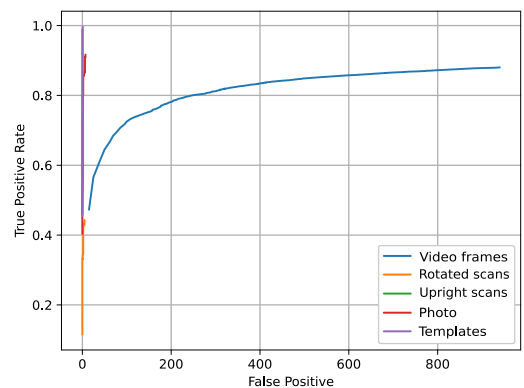
*b)*

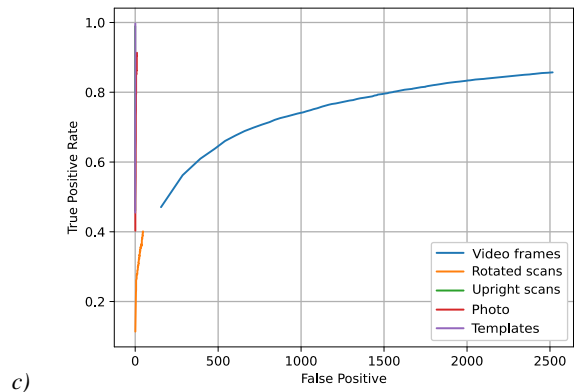

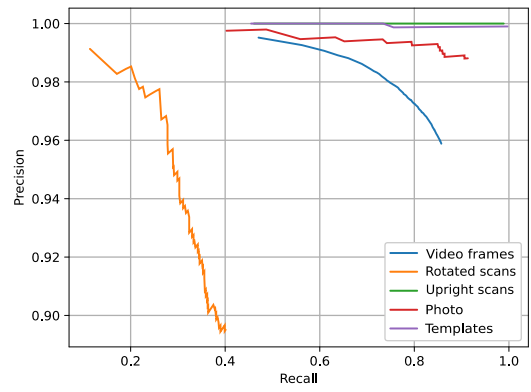

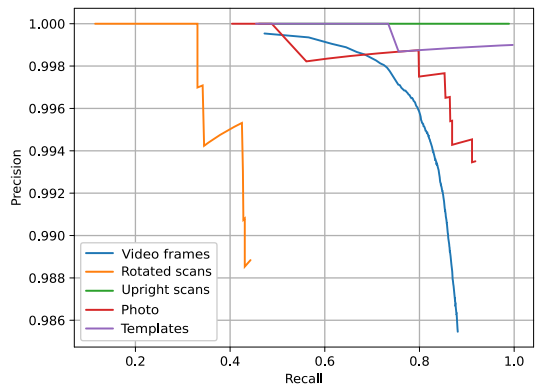

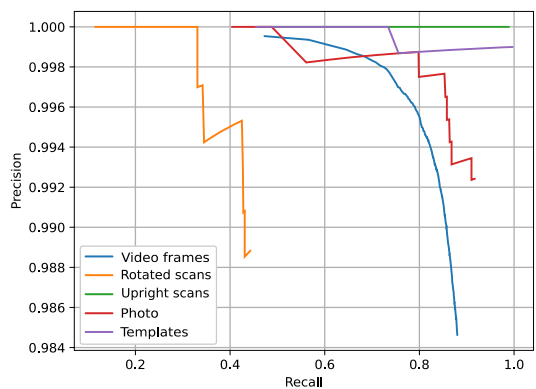

Fig. 10. ROC curves of MTCNN detecting faces of MIDV-2020 with three configurations: a) easy mode with IoU threshold 0.3; b) medium mode with IoU threshold 0.5; c) hard mode with IoU threshold 0.7

Fig. 11. Precision-Recall curves of MTCNN detecting faces of MIDV-2020 with three configurations: a) easy mode with IoU threshold 0.3; b) medium mode with IoU threshold 0.5; c) hard mode with IoU threshold 0.7

From the Figures 10 and Figure 11, we can see that the performance of MTCNN on scan copies of identity documents in the upright orientation («Upright scans» subset, green curve) and templates («Templates» subset, violet curve) are superior to the others, since the identity documents in these two types are always placed properly without any shift, reflect, occlusion etc. and the images have been captured in the perfect condition. There is almost no false detection in these two types. Instead, the face detection on scan copies of rotated documents («Rotated scans» subset, orange curve) shows relatively worse result, which is more pronounced in Figure 11.

We can see that MTCNN performs well for detecting faces on the document images as a baseline model pretrained on general images. Even for some hard cases, MTCNN can still detect the faces on the documents (see Figure 12). Moreover, the performance of MTCNN on MIDV-2020 is even better than MTCNN detecting the faces on general images of WiderFace (i.e. 0.85 on easy set) which is used to train MTCNN. It shows the stability of MTCNN even on the heterogeneous document images. It has to point out that there are two reasons may help MTCNN to obtain such good results. First, the faces presented on the documents are frontal faces no matter how the documents have been placed in space. Comparing to the profile or the faces with extreme pose included in WiderFace, the frontal faces are much easy to detect. Secondly, the ground truth of bounding boxes of faces in MIDV-2020 were semi-automatically annotated with MTCNN, which may also alleviate the detection error of MTCNN. Nevertheless, for the extreme complex cases such as the images with the strong reflect, extreme shooting angle or partial face as shown in Figure 12, the general face detection model still fails to detect the faces. These extreme hard cases have seldom appeared in the general face detection datasets such as FDDB or WiderFace. Thus, MIDV-2020 provides a useful supplementary material to the current face analysis benchmarks enabling the face detection algorithms to be generalized in more different scenarios or domains.

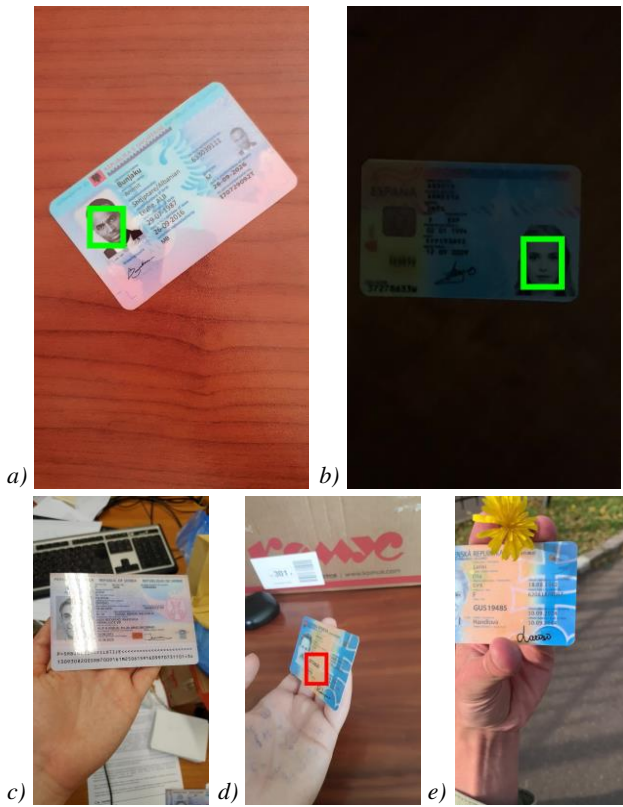

*Fig. 12. Hard cases for the face detection in MIDV-2020; hard cases which were successfully detected using MTCNN: a) high rotation angle (frame alb_id/34/000001) and b) bad illumination (frame esp_id/19/000001); extremely hard cases which MTCNN failed to detect: c) strong reflection (frame srb_passport/91/000181), d) extreme shooting angle (frame svk_id/71/000007), e) partial face (frame svk_id/66/000145)*

## Conclusion

In this article, we presented a new identity documents dataset MIDV-2020, which consists of 2000 scans, 1000 photos, and 1000 video clips of different artificially generated identity documents. In total, the dataset contains 72409 annotated images, which makes it the largest publicly available dataset dedicated to identity documents. We have presented baseline evaluations of the methods of such identity document image analysis problems as document detection, location, and identification; text fields recognition; and face detection. The performed experiments show that the dataset contains examples of images and video clips captured in challenging conditions, which can serve as a motivation for further improvement of document analysis methods and algorithms.

Although the MIDV-2020 dataset is in itself a dataset of fake documents, its usage for developing and evaluating methods of computational document forensics, presentation attacks detection, and other authenticity validation approaches, is limited. As a future work, we plan to expand the dataset to include capturing conditions and document features, which would allow training and evaluation of fraud prevention techniques.

Authors believe that the dataset will serve as a benchmark for document analysis methods and identity document processing systems, and prove valuable for the advancement of the field of document analysis and recognition in general.

## *References*

1. Fang X, Fu X, Xu X. ID card identification system based on image recognition. 12[th] IEEE Conf on Industrial Electronics and Applications (ICIEA), 2017; 1488-1492; DOI: 10.1109/ICIEA.2017.8283074.
2. Attivissimo F, Giaquinto N, Scarpetta M, Spadavecchia M. An automatic reader of identity documents. IEEE Int Conf on Systems, Man and Cybernetics (SMC), 2019; 3525-3530; DOI: 10.1109/SMC.2019.8914438.
3. Kuklinski T, Monk B. The use of ID reader-authenticators in secure access control and credentialing. IEEE Conf on Technologies for Homeland Security, 2008; 246-251; DOI: 10.1109/THS.2008.4534458.
4. Soares A, das Neves Junior R, Bezerra B. BID Dataset: a challenge dataset for document processing tasks. Anais Estendidos do XXXIII Conf on Graphics, Patterns and Images, 2020; 143-146; DOI:10.5753/sibgrapi.est.2020.12997.
5. Ngoc MOV, Fabrizio J, Geraud T. Saliency-based detection of identity documents captured by smartphones. 12[th] IAPR Int Workshop on Document Analysis Systems (DAS), 2018; 387-392; DOI: 10.1109/DAS.2018.17.
6. Arlazarov VV, Bulatov K, Chernov T, Arlazarov VL. MIDV-500: a dataset for identity document analysis and recognition on mobile devices in video stream. Computer optics, 2019; 43; 818-824; DOI: 10.18287/2412-6179-2019-43-5-818-824.
7. Bulatov K, Matalov D, Arlazarov VV. MIDV-2019: challenges of the modern mobile-based document OCR. 12[th] Int Conf on Machine Vision (ICMV); 2020; 114332N; 717-722; DOI: 10.1117/12.2558438.
8. Chazalon J, Gomez-Kramer P, Burie J-C, Coustaty M, Eskenazi S, Luqman M, Nayef N, Rusinol M, Sidere N, Ogier J. SmartDoc 2017 Video Capture: mobile document acquisition in video mode. 14[th] IAPR Int Conf on Document Analysis and Recognition (ICDAR), 2017; 04; 11-16; DOI: 10.1109/ICDAR.2017.306.
9. Skoryukina N, Arlazarov VV, Nikolaev D. Fast method of ID documents location and type identification for mobile and server application. Int Conf on Document Analysis and Recognition (ICDAR), 2019; 850-857; DOI: 10.1109/ICDAR.2019.00141.
10. Buonanno A, Nogarotto A, Cacace G, di Gennaro G, Palmieri FAN, Valenti M, Graditi G. Bayesian feature fusion using factor graph in reduced normal form. Applied sciences, 2021; 11(4); 1934; DOI: 10.3390/app11041934.
11. Skoryukina N, Faradjev I, Bulatov K, Arlazarov VV. Impact of geometrical restrictions in RANSAC sampling on the ID document classification. 12[th] Int Conf on Machine Vision (ICMV), 2020; 1143306; DOI: 10.1117/12.2559306.
12. Lynchenko A, Sheshkus A, Arlazarov VL. Document image recognition algorithm based on similarity metric robust to projective distortions for mobile devices. 11[th] Int Conf on Machine Vision (ICMV), 2019; 110411K; DOI: 10.1117/12.2523152.
13. das Neves Junior RB, Lima E, Bezerra BL, Zanchettin C, Toselli AH. HU-PageScan: a fully convolutional neural

network for document page crop. IET Image Processing, 2020; 14; 3890-3898; DOI: 10.1049/iet-ipr.2020.0532.
14. Sheshkus A, Nikolaev D, Arlazarov VL. Houghencoder: neural network architecture for document image semantic segmentation. IEEE Int Conf on Image Processing (ICIP), 2020; 1946-1950; DOI: 10.1109/ICIP40778.2020.9191182.
15. Bakkali S, Luqman MM, Ming Z, Burie J. Face detection in camera captured images of identity documents under challenging conditions. Int Conf on Document Analysis and Recognition Workshops (ICDARW), 2019; 55-60; DOI: 10.1109/ICDARW.2019.30065.
16. Chernyshova YS, Sheshkus AV, Arlazarov VV. Two-step CNN framework for text line recognition in camera-captured images. IEEE Access, 2020; 8; 32587-32600; DOI: 10.1109/ACCESS.2020.2974051.
17. Petrova O, Bulatov K, Arlazarov VV, Arlazarov VL. Weighted combination of per-frame recognition results for text recognition in a video stream. Computer optics, 2021; 45(1); 77-89; DOI: 10.18287/2412-6179-CO-795.
18. Bulatov KB. A method to reduce errors of string recognition based on combination of several recognition results with per-character alternatives. Bulletin of the South Ural State University, Series: Mathematical Modelling, Programming and Computer Software, 2019; 12(3); 74-88; DOI: 10.14529/mmp190307.
19. Bulatov K, Razumnyi N, Arlazarov VV. On optimal stopping strategies for text recognition in a video stream as an application of a monotone sequential decision model. Int J on Document Analysis and Recognition, 2019; 22(3); 303-314; DOI: 10.1007/s10032-019-00333-0.
20. Bulatov K, Fedotova N, Arlazarov VV. Fast approximate modelling of the next combination result for stopping the text recognition in a video. 25[th] Int Conf on Pattern Recognition (ICPR), 2021; 239-246; DOI: 10.1109/ICPR48806.2021.9412574.
21. Aliev MA, Kunina IA, Kazbekov AV, Arlazarov VL. Algorithm for choosing the best frame in a video stream in the task of identity document recognition. Computer optics, 2021; 45(1); DOI: 10.18287/2412-6179-CO-811.
22. Chernov TS, Ilyuhin SA, Arlazarov VV. Application of dynamic saliency maps to video stream recognition systems with image quality assessment. Int Conf on Machine Vision (ICMV), 2019; 110410T; DOI: 10.1117/12.2522768.
23. Myasnikov E, Savchenko A. Detection of sensitive textual information in user photo albums on mobile devices. Int Multi-Conf on Engineering, Computer and Information Sciences (SIBIRCON), 2019; 0384-0390; DOI: 10.1109/SIBIRCON48586.2019.8958325.
24. Castelblanco A, Solano J, Lopez C, Rivera E, Tengana L, Ochoa M. Machine learning techniques for identity document verification in uncontrolled environments: a case study. Pattern recognition, 2020; 271-281; DOI: 10.1007/978-3-030-49076-8_26.
25. Council of the European Union. PRADO – Public Register of Authentic identity and travel Documents Online, Source: https://www.consilium.europa.eu/prado.
26. Wikipedia. Category: Serbian masculine given names. Source: https://en.wikipedia.org/wiki/Category:Serbian_masculine_given_names.
27. Fantasy name generators: Azerbaijani names. Source: https://www.fantasynamegenerators.com/azerbaijani-names.php.
28. Generated Photos. Source: https://generated.photos.
29. Karras T, Laine S, Aila T. A style-based generator architecture for generative adversarial networks. IEEE/CVF Conf on Computer Vision and Pattern Recognition (CVPR), 2019; 4396-4405; DOI: 10.1109/CVPR.2019.00453.
30. Dutta A, Zisserman A. The VIA annotation software for images, audio and video. Proc of the 27[th] ACM Int Conf on Multimedia (MM'19), 2019; 2276-2279; DOI: 10.1145/3343031.3350535.
31. VGG Image Annotator (VIA). Source: https://www.robots.ox.ac.uk/~vgg/software/via.
32. Javed K, Shafait F. Real-time document localization in natural images by recursive application of a CNN. 14[th] IAPR Int Conf on Document Analysis and Recognition (ICDAR), 2017; 105-110; DOI: 10.1109/ICDAR.2017.26.
33. Zhu A, Zhang C, Zhi L, Xiong S. Coarse-to-fine document localization in natural scene images with regional attention and recursive corner refinement. Int J on Document Analysis and Recognition, 2019; 22; 351-360; DOI: 10.1007/s10032-019-00341-0.
34. Skoryukina N, Nikolaev DP, Sheshkus A, Polevoy D. Real time rectangular document detection on mobile devices. 7[th] Int Conf on Machine Vision (ICMV), 2015; 94452A; DOI: 10.1117/12.2181377.
35. Tropin DV, Ershov AM, Nikolaev DP, Arlazarov VV. Advanced Hough-based method for on-device document localization. Preprint, Source: https://arxiv.org/abs/2106.09987.
36. Tropin DV, Ilyuhin SA, Nikolaev DP, Arlazarov VV. Approach for document detection by contrours and contrasts. 25[th] Int Conf on Pattern Recognition (ICPR), 2021; 9689-9695. DOI: 10.1109/ICPR48806.2021.9413271.
37. Tropin DV, Konovalenko IA, Skoryukina NS, Nikolaev DP, Arlazarov VV. Improved algorithm of ID card detection by a priori knowledge of the document aspect ratio. Int Conf on Machine Vision (ICMV), 2021; 116051F; DOI: 10.1117/12.2587029.
38. Ngoc MOV, Fabrizio J, Geraud T. Document detection in videos captured by smartphones using a saliency-based method. Int Conf on Document Analysis and Recognition Workshops (ICDARW), 2019; 19-24; DOI: 10.1109/ICDARW.2019.30059.
39. Burie J et al. ICDAR2015 competition on smartphone document capture and OCR (SmartDoc). 13[th] Int Conf on Document Analysis and Recognition (ICDAR), 2015; 1161-1165. DOI: 10.1109/ICDAR.2015.7333943.
40. Liu L, Wang Z, Qiu T, Chen Q, Lu Y, Suen CY. Document image classification: Progress over two decades. Neurocomputing, 2021; 453; 223-240; DOI: 10.1016/j.neucom.2021.04.114.
41. Chiron G, Ghanmi N, Awal AM. ID documents matching and localization with multi-hypothesis constraints. 25[th] Int Conf on Pattern Recognition (ICPR), 2021; 3644-3651; DOI: 10.1109/ICPR48806.2021.9412437.
42. Awal AM, Ghanmi N, Sicre R, Furon T. Complex document classification and localization application on identity document images. 14[th] IAPR Int Conf on Document Analysis and Recognition (ICDAR), 2017; 426-431; DOI: 10.1109/ICDAR.2017.77.
43. Bay H, Ess A, Tuytelaars T, Gool LV. Speeded-Up Robust Features (SURF). Computer Vision and Image Understanding, 2008; 110(3); 346-359; DOI: 10.1016/j.cviu.2007.09.014.
44. Suarez I, Sfeir G, Buenaposada JM, Baumela L. BEBLID: boosted efficient binary local image descriptor. Pattern recognition letters, 2020; 133; 366-372; DOI: 10.1016/j.patrec.2020.04.005.
45. Smith R. An overview of the Tesseract OCR engine. 9[th] Int Conf on Document Analysis and Recognition (ICDAR), 2007; 629-633; DOI: 10.1109/ICDAR.2007.4376991.

46. Smith R, Podobny Z, etc al. Tesseract OCR. Source: https://github.com/tesseract-ocr/tesseract.
47. ABBYY FineReader PDF: the smarter PDF solution. Source: https://pdf.abbyy.com.
48. Jain V, Learned-Miller E. FDDB: A benchmark for face detection in unconstrained settings. University of Massachusetts, Amherst, 2010; UM-CS-2010-009; Source: http://vis-www.cs.umass.edu/fddb.
49. Yang S, Luo P, Loy CC, Tang X. WIDER FACE: a face detection benchmark. IEEE Conf on Computer Vision and Pattern Recognition (CVPR), 2016; 5525-5533; DOI: 10.1109/CVPR.2016.596.
50. Hu P, Ramanan D. Finding tiny faces. IEEE Conf on Computer Vision and Pattern Recognition (CVPR), 2017; 1522-1530. DOI: 10.1109/CVPR.2017.166.
51. Najibi M, Samangouei, Chellappa R, Davis LS. SSH: single stage headless face detector. IEEE Int Conf on Computer Vision (ICCV), 2017; 4885-4894; DOI: 10.1109/ICCV.2017.522.
52. Yang S, Luo P, Loy C, Tang X. From facial parts responses to face detection: a deep learning approach. IEEE Int Conf on Computer Vision (ICCV), 2015; 3676-3684; DOI: 0.1109/ICCV.2015.419.
53. Zhu Y, Cai H, Zhang S, Wang C, Xiong Y. TinaFace: strong but simple baseline for face detection. Preprint, Source: https://arxiv.org/abs/2011.13183.
54. Li H, Lin Z, Shen X, Brandt J, Hua G. A convolutional neural network cascade for face detection. IEEE Conf on Computer Vision and Pattern Recognition (CVPR), 2015; 5325-5334; DOI: 10.1109/CVPR.2015.7299170.
55. Zhang K, Zhang Z, Li Z, Qiao Y. Joint face detection and alignment using multitask cascaded convolutional networks. IEEE Signal Processing Letters, 2016; 23(10); 1499-1503; DOI: 10.1109/LSP.2016.2603342.
56. Shi X, Shan S, Kan M, Wu S, Chen X. Real-time rotation-invariant face detection with progressive calibration networks. IEEE Conf on Computer Vision and Pattern Recognition (CVPR), 2018; 2295-2303.
57. Liu W, Anguelov D, Erhan D, Szegedy C, Reed S, Fu C-Y, Berg AC. SSD: single shot multibox detector. ECCV 2016, Lecture Notes in Computer Science, 2016; 9905; DOI: 10.1007/978-3-319-46448-0_2.
58. Zhang S, Zhu X, Lei Z, Shi H, Wang X, Li SZ. S^3FD: single shot scale-invariant face detector. IEEE Int Conf on Computer Vision (ICCV), 2017; 192-201; DOI: 0.1109/ICCV.2017.30.
59. Deng Z, Guo J, Zhou Y, Yu J, Kotsia I, Zafeiriou S. RetinaFace: single-stage dense face localisation in the wild. IEEE Conf on Computer Vision and Pattern Recognition (CVPR), 2020; 5203-5212.
60. Lin TY, Maire M, Belongie S, Hays J, Perona P, Ramanan D, Dollar P, Zitnick CL. Microsoft COCO: common objects in context. ECCV 2014, Lecture Notes in Computer Science, 2014; 8693; DOI: 10.1007/978-3-319-10602-1_48.

*Appendix*

*Table 1. Specification of MIDV-2020 fields according to the field groups*

| Document type | Latin alphabet | Numerical fields | Local alphabets | MRZ |
|---|---|---|---|---|
| alb_id | authority; gender; id_number; nationality | birth_date; issue_date; expiry_date; numbers | birth_place; name; surname | mrz_line0; mrz_line1 |
| aze_passport | type; code; number; surname_eng; name_eng; id_number; gender; authority_eng | birth_date; issue_date; expiry_date; expiry_date_2 | surname; name; nationality/nationality_eng; birth_place; authority | |
| esp_id | id_number; gender; nationality; number; name_code | birth_date; expiry_date; issue_date | name; surname; surname_second | |
| est_id | name; gender; nationality; number; number2; number3 | birth_date; id_number; expiry_date | surname | |
| fin_id | gender; nationality | number; birth_date; issue_date; expiry_date; birth_date_2; birth_date_22 | surname; name; code | |
| grc_passport | birth_date; birth_country; expiry_date; gender; issue_date; name_eng; surname_eng; type; number | height | birth_place; name; surname | |
| lva_passport | code; gender; nationality; number; type | birth_date; expiry_date; height; id_number; issue_date | authority_line0; authority_line1; birth_place; name; surname | |
| rus_internalpassport | | birth_date; number | birth_place_line0; birth_place_line1; birth_place_line2; gender; name; patronymic; surname | |
| srb_passport | code; nationality; type | birth_date; expiry_date; id_number; issue_date; number; expiry_date2 | authority; birth_place; gender; name; residence_line0; residence_line1; surname | |
| svk_id | gender; nationality; number | id_number; birth_date; expiry_date; issue_date | name; surname; issue_place | |